# Pedestrian Models for Autonomous Driving Part I: Low-Level Models, from Sensing to Tracking

Fanta Camara[1,2], Nicola Bellotto[2], Serhan Cosar[3], Dimitris Nathanael[4], Matthias Althoff[5], Jingyuan Wu[6], Johannes Ruenz[6], André Dietrich[7] and Charles Fox[1,2,8]

*Abstract*—Autonomous vehicles (AVs) must share space with pedestrians, both in carriageway cases such as cars at pedestrian crossings and off-carriageway cases such as delivery vehicles navigating through crowds on pedestrianized high-streets. Unlike static obstacles, pedestrians are active agents with complex, interactive motions. Planning AV actions in the presence of pedestrians thus requires modelling of their probable future behaviour as well as detecting and tracking them. This narrative review article is Part I of a pair, together surveying the current technology stack involved in this process, organising recent research into a hierarchical taxonomy ranging from low-level image detection to high-level psychology models, from the perspective of an AV designer. This self-contained Part I covers the lower levels of this stack, from sensing, through detection and recognition, up to tracking of pedestrians. Technologies at these levels are found to be mature and available as foundations for use in high-level systems, such as behaviour modelling, prediction and interaction control.

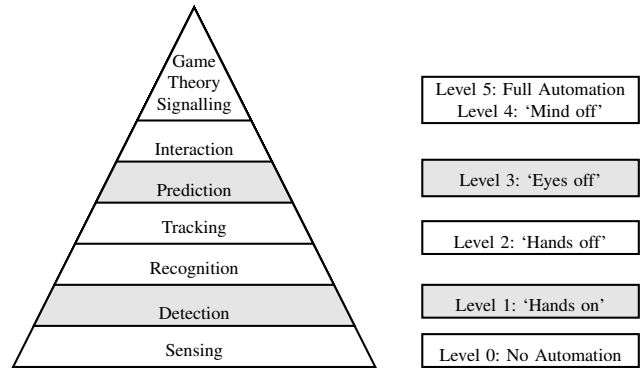

Fig. 1. Main structure of the review.

## I. Introduction

Many organisations are vigorously developing autonomous vehicles (AVs). The technology for vehicles moving in static environments – localising, mapping, planning, and controlling – is well developed [219] and is now available as open-source software [116]. However, in real-world driving environments, human drivers regularly make decisions involving social decision-making that are harder to automate. Autonomous vehicles need additional social intelligence to operate in these complex social environments.

Interacting with pedestrians is a particular type of social intelligence. Autonomous vehicles will need to utilize many different models of pedestrians, each addressing different aspects of perception and intelligence from low-level machine vision detection to high-level psychological and social reasoning. Each of these models can be based on empirical science results or obtained via machine learning. So far, the required models have typically been developed by different research communities, so their integration is currently premature.

At the lower levels of the technology stack, pedestrian modelling requires perceptual methods to detect pedestrians, track their positions and velocities over time, and predict their movements to avoid colliding with them. These methods mostly originate from computer vision and robotics.

At the higher-levels, as researched by psychologists and taught in advanced driver training programmes, drivers may infer the personality of other humans, predict their likely behaviours, and interact with them to communicate mutual intentions. At the higher levels, researchers infer psychological information from perceptual information, for example recognizing pedestrian body language, gestures, and demographics information, to better predict their likely goals and behaviours. Despite the importance of bridging the research between the higher and lower levels, their connection is still thin, both conceptually and in terms of implementations.

A promising method to bridge the higher and the lower levels is probability theory, providing possibilities for quantitative computational interfaces: for example, a pedestrian detector can pass a detection probability to a gesture recognizer, which computes probabilities of particular gestures based on this information, which in turn can be passed to a psychological or game-theoretic behaviour predictor, before the information is finally used to probabilistically compute optimal steering

This project has received funding from EU H2020 interACT (723395).
[1] Institute for Transport Studies (ITS), University of Leeds, UK
[2] Lincoln Centre for Autonomous Systems, University of Lincoln, UK
[3] Institute of Engineering Sciences, De Montfort University, UK
[4] School of Mechanical Engineering, Nat. Tech. University of Athens
[5] Department of Computer Science, Technische Universität München
[6] Robert Bosch GmbH, Germany
[7] Technische Universität München (TUM), Germany
[8] Ibex Automation Ltd, UK
Manuscript received: 2019-03-11; Revisions:2019-10-21, 2020-02-25. Accepted: 26-06-2020.

TABLE I
PROPOSED MAPPING FROM SAE LEVELS TO PEDESTRIAN MODEL REQUIREMENTS.

| SAE LEVEL | DESCRIPTION | MODEL REQUIREMENTS | SECTION |
|---|---|---|---|
| 0 | No Automation. Automated system issues warnings and may momentarily intervene, but has no sustained vehicle control. | Sensing | Sec. II |
| 1 | Hands on. The driver and the automated system share control of the vehicle. For example, adaptive cruise control (ACC), where the driver controls steering and the automated system controls speed. The driver must be ready to resume full control when needed. | +Detection | Sec. III |
| 2 | Hands off. The automated system takes full control of the vehicle (steering and speed). The driver must monitor and be prepared to intervene immediately. Occasional contact between hand and wheel is often mandatory to confirm that the driver is ready to intervene. | +Recognition<br>+Tracking | Sec. IV<br>Sec. V |
| 3 | Eyes off. Driver can safely turn attention away from the driving tasks, e.g. use a phone or watch a movie. Vehicle will handle situations that call for an immediate response, like emergency braking. The driver must still be prepared to intervene within some limited time. | +Unobstructed Walking Models, Known Goals<br>+Behaviour Prediction, Known Goals<br>+Behaviour Prediction, Unknown Goals | Part II Sec. II-A<br>Part II Sec. II-B<br>Part II Sec. II-C |
| 4 | Mind off. No driver attention is required for safety, except in limited spatial areas or special circumstances. Outside of these areas or circumstances, the vehicle must be able to safely abort or transfer control to the human. | +Event/Activity Models<br>+Effects of Class on Trajectory<br>+Pedestrian Interaction Models<br>+Game Theory and Signalling Models | Part II Sec. II-D<br>Part II Sec. II-E<br>Part II Sec. III<br>Part II Sec. IV |
| 5 | Full automation. No human intervention is required at all. | +Extreme Robustness and Reliability | |

Note: '+X' means that 'X' is required in addition to the requirements of the previous level.

and speed values. Such a unified probabilistic stack requires models at all levels to realise quantitative, probabilistic inferences and predictions. Besides surveying the required building blocks, we also examine the maturity of each required level.

Many papers have been published presenting pedestrian models at various levels, but no unifying theory to connect them has yet been produced. The present study is Part I of a linked pair which together survey and unify the stack of required skills from engineered low-level aspects up to high-level aspects involving social decision-making. This Part I reviews the lower-level parts of the stack from sensing, through detection and recognition, to tracking, which together create the required inputs for higher-level AI systems to control interactions reviewed in Part II [28].

Together, these two reviews contribute steps towards such a theory by bringing together, and organising into a new taxonomy (presented via the structure of the papers), research from different fields, including machine vision, robotics, data science, psychology and game theory. We suggest how models from these fields could be linked together into a single technology stack by probability theory. We support this goal by summarizing methods for translating qualitative concepts into simple quantitative statistical models.

Fig. 1 provides an overview of the main structure of the review and links the structure to five levels of driving automation defined by the Society of Automotive Engineers (SAE), ranging from simple driver assistance tools to full self-driving [190]. In our taxonomy, we approximately map requirements for pedestrian modelling to each of these levels, with requirements increasing as levels increase. Table I gives an overview of SAE levels and requirements mapping.

To reach level 0, no automation is required, but some basic sensing is needed to inform the human driver. Very simple sensors can be used, such as the ultrasonic reverse parking sensors currently available commercially, together with very basic signal processing such as distance thresholds causing an audible signal. More complex concepts from our reviews *may* also be added to inform the driver of higher-level information, such as the identity of the particular pedestrian they are about to hit, but this is not *necessary* to *reach* level 0.

To reach level 1, the AV needs to provide driving assistance tools, such as lane keeping and adaptive cruise control (ACC). To do this, it needs to *detect* the road structure and the surrounding objects to help the driver. The AV needs to detect these objects in order to avoid them, but does not yet need to *recognise* them as specific individuals because this is not necessarily needed for obstacle avoidance.

To reach level 2, the AV and the driver must share the driving task, with the vehicle taking full control of the vehicle at certain times. To take full control, it is not sufficient to only detect objects, but it is also necessary to *recognize* and track them over time in order to make *short-term* predictions of their motion and safely avoid them, possibly often passing control to the human, when these simple predictions do not work.

To reach level 3, drivers can turn their attention away from the driving task, but must be prepared to take control occasionally within a certain time. This requires better prediction of pedestrian motion than level 2 in order to reduce take-over requests to humans. For example, adding concepts of likely routes and destinations to pedestrian models reduces the human take-over requests.

Finally, to reach levels 4 and 5, we believe that the AV must understand the driving task as good as a good human driver. Human drivers use complex psychology of pedestrian behaviour as well as their negotiating and signalling behaviours, so these must be replicated by the AV.

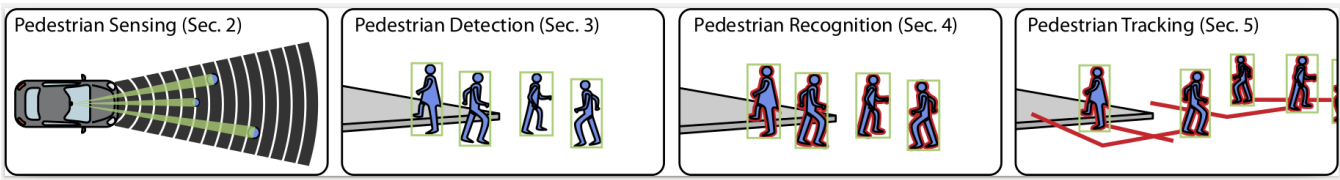

Fig. 2. Structure of the paper.

This Part I begins at the lowest levels of machine vision with sensing (SAE level 0) and detection (SAE level 1), and considers recognition and tracking (SAE level 2) based on them. This Part I is intended to lay the foundations for Part II [28], which then moves up the technology stack to consider SAE levels 3-5. Part II also reviews data sources and other experimental resources useful for building and testing models at all levels.

*Pedestrians* are here defined as humans moving on and near public highways including roads and pedestrianised areas, who walk using their own locomotive power. This excludes, for example, humans moving on cycles, wheelchairs and other mobility devices, skates and skateboards, or those transported by other humans. This review does not cover interactions of traffic participants without pedestrians: a survey on trajectory prediction of on-road vehicles is provided in [133] and a survey on vision-based trajectory learning is provided in [154].

The organization of the review serves as a new taxonomy from relatively well understood quantitative engineering methods at the lower levels, towards less clear qualitative psychological theories of behaviour and interaction. It summarizes some progress in translating these qualitative concepts into simple quantitative statistical models, and identifies a strong need for this process towards quantifying psychological, social, group and interactive models into algorithms for real-world AV control. Each section has an introduction and discussion, which should be readable by researchers from other, especially neighbouring, fields who would like to get an overview of the state of the art and consider how their own field could connect both conceptually and computationally to it. Statistics on included papers are shown in the supplementary material Sect. I. The remainder of this Part I is organized as shown in Fig. 2.

## II. PEDESTRIAN SENSING

Any pedestrian modelling system must begin by collecting sensor data about pedestrians. Detection, tracking and higher-level models may all depend on what information is present at this low-level, so a brief review is provided here. More details on automotive sensors are available in [85]. We classify our review into passive and active sensors. Active sensors actively send pulses into the environment that are reflected and detected while passive sensors detect physical phenomena already present in the environment. A summary of common AV sensors with their range and accuracy is provided in Table II.

### A. Passive Sensors

*a) Manual Detection and Labelling:* The most basic method of sensing pedestrians is to use human perception, which is often used in offline studies, such as for conducting on-street surveys or annotate recordings of such surveys made with other sensors [29], [30]. Humans still have advantages over automated systems since they can use their full intelligence to subjectively annotate otherwise difficult events, such as the meanings of body language, emotions, and gestures. In particular, manual detection of pedestrians is needed and used as ground truth data for machine learning algorithms as in [247] where human experts were asked to detect people as a baseline for a comparison against machine algorithms.

*b) Video Cameras:* One of the most commonly used sensors is the video camera, because it is cheap and easy to install. For example, [75] proposed a survey and experiments on pedestrian detection using monocular cameras. In [252] the shadow of moving objects is removed from the foreground images in order to improve the accuracy of the detection. In [107], shadows are automatically removed from the images in HSV color space. On the contrary, Wang and Yagi [226] treated shadow as helpful information for their appearance-based pedestrian detector.

*c) Stereo Pair Video Cameras:* Traditionally, 3D machine vision was a less-developed research field than 2D image processing [102]. It uses two (or more) images from cameras, placed some distance apart, to estimate the stereo disparity between them and, ultimately, the distance in 3D space. Disparity describes the difference in location of corresponding features seen by the left and right cameras [212, ch. 11]. Disparity estimation methods fall into two classes: pixel-based methods (similar to optical flow), which estimates disparity at each pixel based on colour similarity to its neighbours; feature-based methods, which find a smaller number of statistically *interesting* points in the image (such as corners) and compute only their disparities. In recent years, these algorithms have become standard and very fast hardware implementations have enabled both real-time use and integration into consumer-style camera products [112]. Hence, it is now possible to consider a stereo camera as a single device at the sensor level for detecting humans. For example, in [117], pedestrians are detected using dense (i.e. pixel-based) stereo camera images. Ess *et al.* [76], instead, implemented a stereo vision-based detection algorithm that extracts visual features and performs pedestrian detection from a mobile platform.

*d) Passive Infrared Imaging:* Pedestrians' bodies radiate heat in the infrared (IR) spectrum, which may be easier to detect than the visible one. For example, Xu *et al.* [82]

developed a pedestrian detection and tracking method using a night-vision camera. [209] proposed a pedestrian detection method using infrared images. Cielniak *et al.* [48] presented a technique that combines color and thermal vision sensors data to track multiple people. Unlike visible light, IR does not allow to distinguish a single body from a group of pedestrians, but this technology can be useful for detecting and identifying objects in foggy conditions [143].

*e) Passive Ultrasonic Sensors:* When a moving object enters and then leaves the detection area, the sound energy increases and then decreases: the role of a passive ultrasonic sensor is to measure the produced acoustic energy [72]. This technique is not very reliable, as it might not be able to detect single moving objects from groups, and it is also dependent on weather conditions.

*f) Piezoelectric Sensors:* A review on tactile sensor detection of humans is provided in [218]. Piezoelectric sensors generate an electric impulse on touch contact, such as pedestrians stepping onto a sensed ground region, or making contact with an AV itself. This can become very expensive because it requires the installation of many piezoelectric sensors in the study area, for instance on the floor of the pedestrian infrastructure. It is useful as a last-resort sensor to detect actual collisions when other sensors have failed. In some limited (small but very high density) environments, it may be useful to monitor pedestrian movements around a sensor-filled floor, e.g. in a heavily pedestrianized area shared with last mile robots.

*g) ID Sensors:* These devices are attached to or carried by pedestrians and they transmit unique identifying tags as well as simplifying localisation, and include infrared and RFID (Radio-Frequency IDentification) badges. Schulz *et al.* [198] developed a tracking system which combines ID sensor information with anonymous ones, such as lidar (see Sect. II-B0a), in order to improve tracking accuracy. Versichele *et al.* [223] proposed to use Bluetooth for person tracking based on unique MAC (Media Access Control) addresses emitted continually by many personal devices already carried by pedestrians, such as mobile phones. In [94], camera images are fused with an omnidirectional RFID detection system using a particle filter in order to enable a mobile robot to track people in crowded environments.

### B. Active Sensors

*a) Lidar (Light Imaging Detection And Ranging):* This sensor is mainly used for localisation and detection of traffic participants, such as pedestrians, cars, bicycles, etc. It makes use of laser beams and calculates the distance to obstacles (objects, walls, people) by measuring the time gap between sending and receiving impulses; some lidar have a 360 degrees detection range. It can be used to determine the direction, speed and trajectory of moving objects. For instance, Dewan *et al.* [67] presented a model-free detection and tracking of dynamic objects with 3D lidar data in complex environments. Objects are detected and segmented thanks to multiple motion cues, then their estimated motion model is used for tracking. Arras *et al.* [6] proposed a similar supervised classifier to detect people using a 2D lidar. In this case, AdaBoost (Adaptive

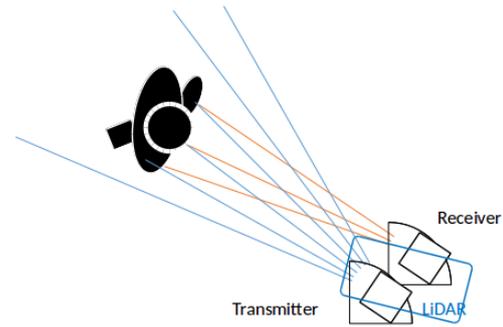

(a) The working principle of a lidar

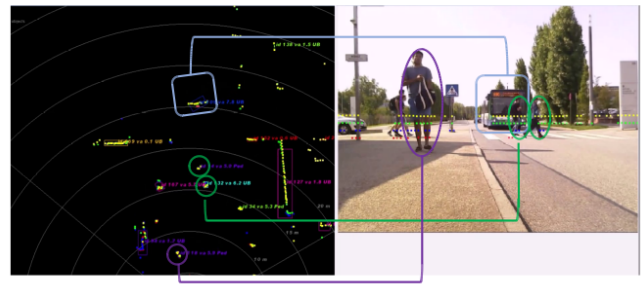

(b) Detection of road users with a 2D lidar

Fig. 3. The working principle of a lidar and its detection of road users.

Boosting), a binary boosting algorithm that combines a set of weak classifiers into a strong classifier, is used to detect features of the laser beams corresponding to peoples' legs in different environments. Gonzalez *et al.* [97] combined lidar and RGB camera data for pedestrian detection. Lidars can be used in any weather conditions, but they can be quite expensive, especially when a range of more than 30m is needed [14]. Fig. 3(a) shows the working principle of lidar and Fig. 3(b) shows the detection of road users using a lidar.

*b) Radar (Radio Detection And Ranging):* This sensor was first used during World War II. Radars emit a radiation from their antenna, which receives back the radiation reflected by passing objects. There are two types of radar: one which transmits a continuous wave of constant frequency to determine the speed of moving objects based on the Doppler principle, where objects with no relative motion are not detected [122]. The second type, frequency modulated continuous wave (FMCW), transmits a continuous changing frequency, which can detect static and moving objects [46].

*c) Active Infrared Sensors:* These sensors are composed of a transmitter that emits infrared light, a receiver that captures the reflected light, and a data collection unit that measures the time of flight of the emitted infrared light. Objects' speed can be detected by sending over two or more beams of infrared light. Their range varies from a few to tens of meters. The Kinect sensor [249], a popular RGBD (red, green, blue, depth) camera, is a particular example of an active infrared sensor. It uses a complex known pattern of thousands of rays and measures their movement in the reflected image to infer distance, similarly to a lidar. A review of computer vision techniques based on the Kinect sensor is proposed in [101].

TABLE II
RANGE AND ACCURACY FOR COMMON AV SENSORS.

| SENSOR | RANGE | ACCURACY |
| --- | --- | --- |
| STEREO CAMERAS | From 0.5m up to several tens of meters [19] | Disparity error of 1/10 pixel (correspond to about 1m distance error if the object is 100m far away) [169] |
| INFRARED | From a few cm to several meters [85], [108] | Temperature accuracy of +/-1°C, can measure temperatures up to 3,000°C [85] |
| ULTRASONIC | From 2cm to 500cm [36], [195] | About 0.3mm [36], [195] |
| RFID | Several meters [256], [84] | A few centimeters [256], [84] |
| LIDAR | Up to 300m [251], [193] | Up to 2cm [62], [193] |
| RADAR | • Short range: 40m, angle 130° [160], [103], [100]<br>• Middle range: 70m to 100m, angle 90° [160], [103]<br>• Long range automotive radar from less than 1m to up to 300m (opening angle up to +/-30°, a relative velocity range of up to +/-260km/h) [62], [160], [204] | • Short range: Less than 0.15m or 1% [160], [103], [100]<br>• Middle range: Less than 0.3m or 1% [160], [103]<br>• Long range: 0.1m e.g. Bosch LRR3 77 GHz, range 250m [62] |

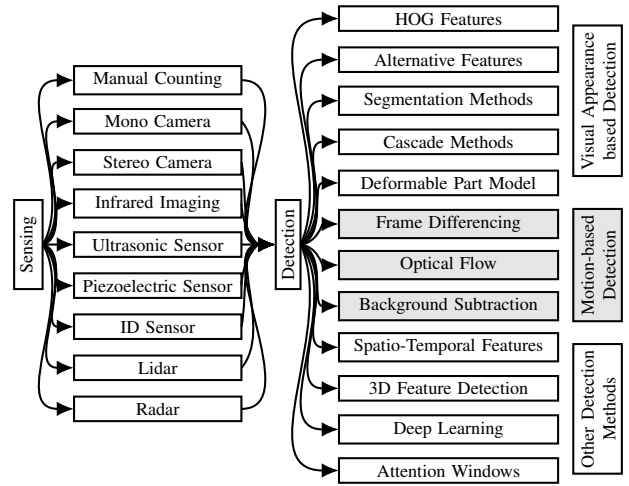

Fig. 4. Pedestrian sensing and detection techniques.

*d) Active Ultrasonic Sensors:* They emit sound waves and a detector senses the sound waves reflected by passing objects. This low-cost sensing method is immune to lighting conditions and does not require significant maintenance. However, it can be seriously affected by weather conditions and it is typically not accurate enough in certain areas [36].

### C. Discussions

Most autonomous vehicles today are using a mix of lidar, radar, and stereo vision. Visual RGB images are most commonly used as the base for detection, and feature-based localisation and mapping. Lidar or radar provide more reliable, but more expensive sensing capabilities for safety-critical aspects such as collision avoidance. While stereo cameras and radar are already used in commercially-available vehicles – for example in lane departure and adaptive cruise control systems, respectively – we expect that lidars will be used as well due to expected drops in prices. In recent years, lidar has been the main source of point cloud localisation and mapping in high-precision sensing for research work, but developments in millimeter radar and stereo cameras are making them increasingly competitive for this purpose. Manual annotation of image data remains necessary for recognition of difficult detailed features such as pedestrian eye contact and body language meanings, but for other tasks even including the creation of training sets for machine learning, is now replaced by automated methods, including semi-supervised approaches which allow quite small manual training sets to be bootstrapped with much larger unannotated data.

## III. PEDESTRIAN DETECTION MODELS

A previous review of pedestrian detection is presented in [71]. Here we summarize some of the key detection methods that are particularly relevant to AVs. Different techniques are used for detection, which can be classified into six main categories: visual appearance-based detection, motion-based detection, spatio-temporal feature detection, 3D feature detection models, deep learning methods and attention-windows detection. In computer vision, the detection problem can be viewed as a special case of image classification: given a candidate image window, the detection seeks to classify the latter as a pedestrian or non-pedestrian. The same concept applies to other types of sensors with their own detection windows. Fig 4 summarizes the sensing technologies and the pedestrian detection techniques described in this section.

### A. Visual Appearance-Based Detection

Unlike motion-based methods, feature-based methods can operate with a single still image, as they look only for static patterns rather than changes over time.

*a) HOG-SVM:* One of the most commonly used pedestrian detectors is based on the combination of HOG (Histogram of Oriented Gradients) and SVM (Support Vector Machine). HOG [60] is a technique that was invented for the purpose of human detection. After training, a classifier can determine whether a proposed HOG corresponds to a pedestrian or not (Fig. 5). The OpenCV vision library [24] has a generic implementation of an object detector based on this method, which can be applied to pedestrian detection.

*b) Alternative Features:* Sometimes used in place of HOG, alternative features including point descriptors, e.g. BRISK (Binary Robust Invariant Scalable Keypoints) and SIFT (Scale Invariant Feature Transform), are used to detect characteristic features of an image, such as corners or edges [192] [20]. Other forms of gradient features and edge detectors [33] are less sensitive to illumination compared to color descriptors. Texture features, such as Local Binary Patterns (LBP), assign a class to each local window. Groups of classes in nearby windows can then be classified as pedestrians or non-pedestrians. For example, [3] proposed a face recognition method based on the LBP feature descriptor. [163] used LBP with spatial pooling for a robust pedestrian detection.

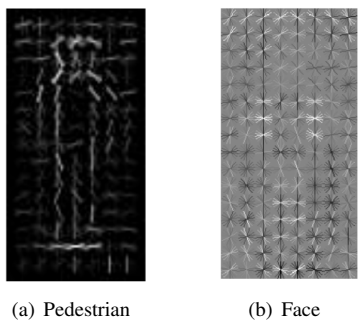

(a) Pedestrian  (b) Face

Fig. 5. Examples of HOG features [60].

*c) Cascade-based Detection:* The detector proposed by [224] is composed of a sequence of classifiers, trained using Haar-like visual features, where each classifier can pass or not a sub-region to the following one. Zhu *et al.* [258] proposed a person detection method using a cascade (40 levels) of HOG-SVM detectors combined with Adaboost for feature selection. In [42], Chen *et al.* developed a person detection approach using a cascade classifier based on Adaboost with rectangle features and edge orientation histogram (EOH) features.

*d) Segmentation Methods:* These include methods such as the Mean-Shift clustering [27], watershed, and grab-cut, which divide the image into regions typically having similar or smoothly changing colour and texture characteristics. These regions can then be tested directly for pedestrians presence through shape, texture and other statistics as in [188], where people were detected and segmented based on a probabilistic method that describes the shapes of their different postures.

*e) Deformable Part Model:* Deformable Part Model (DPM) is a popular detection model. It has been originally proposed for the Pascal VOC challenge for object (including pedestrian) detection and recognition [77]. DPM splits an object into several parts arranged in a deformable configuration and can be used for pedestrian classification as in [79]. This method can deal with significant variations in shape and appearance. A fast implementation of DPM applicable for person detection is proposed in [233].

### B. Motion-based Detection

*a) Frame Differencing:* This method consists in computing the difference between the current frame and a reference one (usually the first frame). In [74], a person detector was developed using optical flow computed on regions selected by frame differencing on camera data recorded from a vehicle. Selected regions are then passed to a wavelet-based features classifier combined with template matching. Park *et al.* [165] proposed an approach that uses coarse-scale optical flow to stabilize camera frames with temporal difference features for pedestrian detection and human pose estimation, and tested on the Caltech pedestrian benchmark [70].

*b) Optical Flow:* This technique assigns a direction and a velocity of motion to each pixel of two consecutive frames, as in [225]. Fernández-Caballero *et al.* [83] used optical flow and frame differencing for human detection on infrared camera images for a security mobile robot platform. Another use of optical flow for detection and tracking is proposed in [67] using 3D lidar data.

*c) Background Subtraction:* This method builds a background model used as a reference model in order to detect moving objects. This modelling is based on the assumption that the background is static. It consists in extracting an estimate of the background from the rest of the image by using some methods such as mean filter, running Gaussian average, etc. Background modelling has two variants: the recursive algorithm, which updates each frame with the estimate of the background, and the non-recursive algorithm, which stores a buffer with the previous frames and the background estimated from them. In [201], Sheikh *et al.* developed a background subtraction model that can detect humans and objects in moving camera images. Their method builds background and foreground appearance models based on the background trajectory estimated by a RANSAC algorithm.

### C. Other Detection Models

*a) Spatio-Temporal Features:* These are commonly used in video codecs, such as Theora and H.264, because they are statistically efficient summary descriptors of natural video. As such, they are also candidates for informative classification features. Oneata *et al.* [162] used these features with a supervoxel method for human detection in videos.

*b) 3D Feature Detection:* These models rely on 3D sensors, such as depth cameras and 3D lidars. Depth information enables more robust detection algorithms. For example, the authors in [234] proposed an online learning method based on a 3D lidar cluster detector, a multi-target tracker, a human classifier and a sample generator. The cluster detection starts by removing the ground plane, then point clusters are extracted from the point clouds using the Euclidean distance in 3D space and finally a human-like volumetric model is fitted to the clusters for filtering. Yan *et al.* [235] took advantage of multiple (2D and 3D) sensor detectors to train an online semi-supervised human classifier for a mobile service robot. A depth-based person detector is presented in [151]. This detector applies template matching on depth images. To reduce the computational load, the detector first runs a ground plane estimation to determine a region of interest, which is the most suitable to detect the upper bodies of a standing or walking person. In [58], a mobile robot equipped with an RGB-D camera is used to detect people. Munaro and Menegatti [156] proposed a real-time detection and tracking system based on RGB-D camera data capable of detecting people within groups or standing near walls.

*c) Attention Windows:* In their basic forms, the classifier-based detection methods above may assume that every possible location and size window of a 2D or 3D image will be tested for pedestrian detection. Such 'sliding windows' can be computationally slow, unless the tests are performed in parallel (e.g. on a GPU) or some form of attention model is used to restrict the search. It is common to use a simple, fast, and inaccurate detector set to have many false positives and few false negatives, to decide whether a window should be explored further or not [200]. In this case, a more advanced but slower method would be applied to test the most interesting windows. Prokhorov [173], for example, developed a road obstacle detector based on attention windows with potential application to pedestrian detection.

*d) Neural networks ('deep learning'):* Neural networks [98] are hierarchical-in-the-parameters regression models which seek to minimise an error function $E$ between $N$ desired vector outputs $c^{(n)}$ for $n \in \{0, N-1\}$ and a function $F$ of input vectors $x^{(n)}$ (including an element which is always 1) with parameters $\theta$,

$$E = \sum_n \|c^{(n)} - F(x^{(n)}; \theta)\|^2, \qquad (1)$$

where $F$ is comprised of layers of 'node' functions,

$$y_j = f(a_j), \quad a_j = \sum_i w_{ji} y_i, \qquad (2)$$

and $f$ is any nonlinear function, $w_{ji} \in \theta$ are weights from any node $i$ in a lower layer to any node $j$ in the layer above it, and $y_i$ for the lowest layer are elements of the input vector $x_i^{(n)}$. The vector formed from $y_l$ for all nodes $l$ in the top layer is the value of $F$. $E$ is then locally minimised by computing *backpropagation* terms $\Delta_i$ for each node,

$$\Delta_i = f'(a_i) \sum_j \Delta_j w_{ji}, \qquad (3)$$

beginning by setting for the top layer nodes $l$,

$$\Delta_l = c_l^{(n)} - F(x^{(n)}; \theta)_l, \qquad (4)$$

then updating the parameters $w_{ji}$ along the direction,

$$-\frac{\delta E}{\delta w_{ji}} = -\Delta_j y_i. \qquad (5)$$

Neural networks date from at least the 1970s [229], but have returned to popularity due to falls in prices of parallel hardware (specifically, graphics cards) which has enabled the use of 'deep' networks having more layers; together with the algorithmic improvements of sharing weights (convolutional neural networks, CNN), pooling [130] and dropout [125] which exploit statistical regularities found in most natural data.

The classifier-based detectors presented so far rely on a two-stage process of feature extraction followed by classification. Neural networks can be used in this way as classifiers given input vectors of features. But increased computing power now enables the raw image to be given directly as input to neural networks having more layers, which can learn their own feature sets in the lower layers, enabling features to be learned, rather than manually chosen, to optimise performance in specific tasks. For example, [5] proposed a real-time pedestrian detector using 'deep network cascades'.

Like other classifier-based detectors, neural networks themselves only learn a mapping from input to output vectors, so to apply them to *detection* of objects in images, some scheme like the attention windows of section III-C0c is needed to propose regions of interest. R-CNN [96] computes region proposals with any non-neural method such as 'selective search'. It computes features for each proposal region using a large CNN, then classifies these features sets using class-specific linear SVMs and also uses linear regression to refine the region from the features. Faster R-CNN [181] extends a CNN with layers for region proposals and layers for classification, using them to propose then classify regions. YOLO [177], [178], [179] similarly extends a CNN with layers for both region proposal and classification, but runs them at the same time with classification based on approximate rather than finally proposed regions. It is able to detect about twenty different classes such as people, cars, bicycles and trucks in real time video. Mask R-CNN [104] finds segmentations as well as rectangular regions, by extending Faster R-CNN with layers predicting masks for regions.

### D. Discussions

Traditionally, a wide variety of image features have been developed by hand and matched with a wide variety of classifiers, to find good performance in pedestrian detection. Until recently, the HOG-SVM method was the best known [16]. Pedestrian detection, like most classification tasks, has however recently been revolutionized by price falls in parallel hardware such as GPUs, which have enabled classical neural network algorithms with small modifications ('deep learning') to outperform hand-crafted methods for the first time. It seems likely that neural network methods will completely replace all others. The same GPU hardware also enables pixel-wise algorithms, such as optical flow, to be massively accelerated. They might not be necessary though if neural networks alone achieve the required performance.

The implementation of a person detection method for an AV is one of the major practical challenges. OpenCV[1] library provides open-source implementation of many computer-vision algorithms (in C++ and Python), mainly aimed at real-time processing. It contains feature extraction methods such as HOG, SIFT, BRISK. It also includes a C++ implementation of DPM. In addition, LibSVM[2] is a popular implementation of SVM classification algorithm. The lidar-based leg detector in [6] is implemented as a Robot Operating System (ROS) module[3]. Again, the ROS implementation of the depth-based detector in [151] is available[4]. In addition, an offline version of the 3D lidar-based approach in [234] is implemented as a ROS module[5]. The authors of the RGBD-based detector in [156] provide the implementation of their algorithm[6]. Many DL-based approaches provide their code for reproducibility and comparison: YOLO[7], R-CNN[8], Faster R-CNN[9] and Mask R-CNN[10].

---

[1] https://opencv.org/
[2] https://www.csie.ntu.edu.tw/~cjlin/libsvm/
[3] https://github.com/wg-perception/people
[4] https://github.com/strands-project/strands_perception_people/
[5] https://github.com/LCAS/FLOBOT
[6] http://pointclouds.org/documentation/tutorials/ground_based_rgbd_people_detection.php

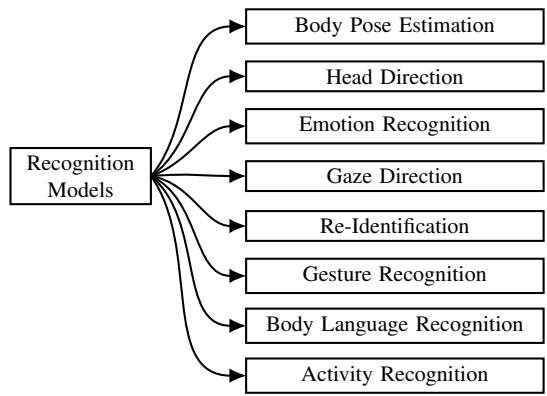

Fig. 6. Pedestrian attributes for recognition models.

High performance of deep learning models comes at a price: they require larger training data (sometimes several millions of examples), longer training times (up to several days), and their computational cost is more important than for simpler detectors [248]. In some cases, DL methods cannot reach real-time performance [5] and are outperformed by simpler methods such as HOG [221].

## IV. PEDESTRIAN RECOGNITION MODELS

While detection refers to finding the presence or absence of pedestrians at locations and scales in images, *recognition* here refers to the recognition of attributes of pedestrians given such detections. Recognition takes as input the localised window of visual or other sensor data forming the detection, and yields as output some information about the particular pedestrian detection. In some cases, this could include their actual identity – identity recognition – but our use of the term here also includes recognition of attributes such as their body pose and facial features. Recognition refers to these *tasks*, while *classification* here refers to processes that perform recognition specifically by mapping inputs into discrete rather than continuous output classes. Figure 6 presents a set of attributes used for pedestrian recognition and a summary of the recognition models and papers reviewed in this paper is given in the supplementary material Sect. II.

### A. Recognition of Body Pose

While full-body tracking is discussed below, some methods may attempt to classify from single images some basic information on pose, such as the head direction of the pedestrian into facing AV/not facing AV. Where the pedestrian body state is known – as resulting from skeleton and other body tracking – it may contain useful information about pedestrians' goals and intentions, which may be extracted by classifiers operating at a higher-level – on the tracked body configurations rather than on the raw images or other sensor data. Cao *et al.* [34], [35] presented OpenPose, a real-time multi-person pose estimation software that uses CNNs to detect people in 2D images and Part Affinity Fields (PAF) is used to associate body parts to the detected people. Shotton *et al.* [203] developed 3D human pose estimation based on body parts representation. Their method relies on depth features, randomized decision trees and forest algorithms for classification, and outputs a proposal position for each detected body part. The method was tested on motion capture and synthetic data.

[7]https://pjreddie.com/darknet/yolo/
[8]https://github.com/rbgirshick/rcnn
[9]https://github.com/rbgirshick/py-faster-rcnn
[10]https://github.com/facebookresearch/Detectron

Iqbal *et al.* [110] proposed a graphical model optimized by a integer linear programming (ILP) to estimate and track multiple people in videos; the used data is made available as a new dataset called PoseTrack. Tompson *et al.* [220] combined a deep CNN with a Markov Random Field to estimate human pose from monocular images. Fragkiadaki *et al.* [89] proposed a method using recurrent neural networks with an Encoder-Recurrent-Decoder (ERD) architecture to predict body joint displacements. ERD is an extension of LSTMs. Martinez *et al.* [144] proposed a method using RNN with Gated Recurrent Unit (GRU) architecture without requiring a spatial encoding layer and allows to train a single model on the whole human body. Tang *et al.* [214] proposed a model that extends the work in [89] and [144]. Their work is based on the observation of human skeleton sequences and uses deep neural networks (Modified High-way Unit (MHU)) to remove motionless joints, estimate next moves and perform human motion transfer. Gosh *et al.* [95] used a Dropout Autoencoder LSTM (DAE-LSTM) to extract structural and temporal dependencies from human skeleton data. Manual annotations are not needed because a tracker gives the actual direction of movement. Kohari *et al.* [123] used a CNN model to estimate human body orientation for a service robot.

### B. Recognition of Head Direction

The primary use of extracting the head direction in pedestrian-AV interaction is epistemological: a pedestrian facing the AV – and/or establishing direct eye contact with it – is a good indicator that the pedestrian has seen the AV and knows it is there, and therefore will be planning their own behaviors on the assumption that they will have to interact with it. In contrast, a pedestrian who has not seen the AV, unless relying on auditory cues, may just step into the road with no idea that a potentially dangerous interaction is about to occur [230] [12]. Darrell *et al.* [61] developed a real-time human tracking and behaviour understanding system, called Pfinder. The system converts human head and hands into a statistical model of color and shape in order to deal with different viewpoints. Schulz and Stiefelhagen [197] estimated pedestrian head pose using multi-classifiers for different monocular grayscale images; depth information within the detection bounding box is also taken into account. Flohr *et al.* [86] proposed a model that can detect pedestrian body and head orientation from grayscale images based on a pictorial structure method.

### C. Recognition of Gaze Direction and Eye Contact

Algorithms for gaze tracking and eye contact detection are not yet robust, and in laboratory eye tracking experiments require expensive precision equipment to be installed on the subjects' heads. Benfold and Reid [17] proposed a method which infers the gaze direction from a head pose detector based on HOG and colour features. The head pose is classified using randomised ferns, i.e., similar to decision trees, and the tracking is done frame-by-frame based on the head detector using multiple point features. Baltrusaitis *et al.* [9] developed the open-source OpenFace, running in real-time with a simple webcam. It is suitable for facial behavior analysis, in particular for facial landmark detections, head pose estimation, facial action recognition and eye-gaze estimation.

### D. Emotion Recognition

Pedestrian emotion recognition might be useful to inform about their crossing intention. For example, an angry pedestrian might be more likely to behave more assertively in crossing the road in front of an AV. Cornejo *et al.* [56], [55] developed a facial expression recognition method that is robust to occlusions. The occluded facial expression is reconstructed with a robust principal component analysis (PCA) method, facial features are extracted using Gabor wavelets and geometric features in [56] and using CENTRIST features in [55], recognition is performed with KNN and SVM as classifiers. Cambria *et al.* [32] proposed a new categorization model for emotion recognition systems and [31] reviewed sentiment analysis methods. Poria *et al.* [171] developed a CNN model with a convolutional recurrent multiple kernel learning that can extract features from multimedia data such as audio, videos, and text. The method has been tested on Youtube videos and ICT-MMMO dataset. Den Uyl and Van Kuilenburg [65] developed the FaceReader, an online facial expression recognition system, which is robust to the head pose, orientation and lighting conditions.

### E. Recognition of Pedestrian Identity for Re-Identification

Person re-identification (re-ID) is the problem of recovering the identify[11] of the same person with different clothing across different images, under different camera views, weather, lighting, and other environmental conditions. Ahmed *et al.* [2] developed a deep convolutional network that solves the re-identification problem by computing a similarity value between two image pairs. Their method has been tested on CUHK01, CUHK03 and VIPER datasets. Zheng *et al.* [253] proposed a person re-identification method based on the Bag-of-Words (BoW) model which extracts Color Names (CN) descriptor features from the input image, a Multiple Assignment (MA) is then used to find neighboring local features and finally TF-IDF finds the number of occurrences of visual words. Their method was tested on the Market1501 dataset. In [254], a CNN model with unlabeled images is used to re-identify people. Li *et al.* [134] proposed a filter pairing neural network (FPNN) model for person re-identification, capable of handling challenging conditions such as occlusions.

[11]Identity here is distinct from 'personal information' as defined by privacy laws such as the EU General Data Protection Regulation (GDPR).

### F. Gesture Recognition

Deliberate gestures are the most obvious form of communication from body pose. For example, a pedestrian may wave a vehicle on to show that they intend to give it priority in a crossing. A previous review on hand gesture recognition is provided in [150] and more recently Rautaray and Agrawal [176] presented a survey for interactions with a computer. Chen *et al.* [40] used a real-time tracker with hidden Markov models (HMM) to recognize hand gestures. Freeman and Roth [90] used orientation histograms for gesture recognition. Their real-time method can recognize about 10 different hand gestures. Ren *et al.* [182] developed a robust hand gesture recognition system for active infrared (Kinect) sensors. Their method is based on template matching for part-hand gesture recognition and a new distance metric called Finger-Earth Mover's Distance (FEMD) is used to measure the similarity between two hand shapes. Other gesture recognition methods based on HMMs are proposed in [23] [132].

### G. Body Language Recognition

In addition to deliberate gestures, unconscious body language, including stance and gait (walking style), may also be a predictor of pedestrian assertiveness in interactions, and of other behaviours. As with gesture recognition, body language recognition relies on recognition of body pose, followed by classification of this pose. Quintero *et al.* [174] proposed a hidden Markov model for pedestrian intention recognition based on 3D positions and joint displacements along the pedestrian body. In [227], a human gait recognition method is proposed, combining background subtraction with PCA for dimensionality reduction. A supervised pattern classification is finally performed to recognize the gait.

### H. Activity Recognition

Pedestrian activity recognition is of particular importance for autonomous vehicles. A lot of work is ongoing for service robots and AVs. A more complete review on human activity recognition methods is proposed in [68]. Chaaraoui *et al.* [37] used contour points of human silhouette to recognize human actions for real-time scenarios. Doll'ar *et al.* [69] used spatio-temporal features for both human and rodent behaviour recognition. Vail *et al.* [222] compared hidden Markov models to conditional random fields for human activity recognition. In [138], a coupled conditional random field is used with RGB and depth sequential information. Coppola *et al.* [54] developed one of the first RGBD-based social activity recognition methods for multiple people. Their method learns spatio-temporal features from skeleton data, which are fused using a probabilistic ensemble of classifiers called Dynamic Bayesian Mixture Model (DBMM).

### I. Discussions

AVs need to recognize pedestrian attributes including pose and possibly identity to help them make more accurate predictions about pedestrians' likely future behaviours. Detection of pedestrians is now mature technology, but recognizing the attributes of these pedestrians within these detections, such as body pose, is a harder and still open research area. Eye direction and eye contact remain particularly difficult as it requires very precise estimation of the positions of small pupils and irises at a long distance. Humans have evolved to be particularly good at recognizing gaze direction for social purposes, but it is hard to replicate. Recognition of emotions may be useful to inform predictions of pedestrians' likely behaviours (e.g. an angry pedestrian may be more likely to push in front of us), and progress has been made in this area in non-real time systems, such as social networks' processing of photographs. But again, recognition from far distances and speeds travelled by AVs for real-time encounters remains challenging and open. It is likely, in the future, that neural network approaches will come to dominate this area as with detection.

Open-source implementations of pedestrian recognition models include Openpose[12] for pose estimation, OpenFace[13] for head pose and eye-gaze estimation and OpenTrack[14] for head tracking. To our knowledge, there is no generally accepted benchmark for pedestrian recognition models. Future research should thus explore the performance and computational efficiency of pedestrian recognition models in the context of autonomous driving.

Recognition of any attribute which enables recovery of a pedestrian's name or other formal identification will fall under data protection laws in most jurisdictions, such as the GDPR across the EU. While re-identification (re-ID) might be particularly useful, for example for use in delivery robots to confirm recipients' identities, the usage of this technology should be carefully assessed with respect to data privacy. The other recognition and tracking algorithms mentioned in this section extract features anonymously, i.e., extracted data does not allow the identification of individuals. Re-ID on the other hand can be used to record and store sensitive personal data, which yields the potential to be misused for public surveillance. For AVs, centralized re-ID might be useful to link individual traffic participants to their previously-observed behavior in traffic enhancing long-term path prediction, but at the cost of severe intrusion into the privacy of road users. This will raise a host of ethical and legal issues when such accuracy is reached by rapidly accelerating machine vision research, such as selling data of individual's locations and behaviours to insurance and advertising companies, or use by local authorities or law enforcement agencies [88].

## V. PEDESTRIAN TRACKING MODELS

Pedestrian tracking is the process of updating the belief about a pedestrian's location from a temporal sequence of data. More specifically, tracking consists in determining the position and possibly orientation or velocity of a given object over time. A pedestrian track is a sequence of their locations over time. A pedestrian pose track is a sequence of a pedestrian's body pose states over time. When multiple pedestrians are present, tracking requires separating the pedestrians from each other and associating the identities of the pedestrians with tracks. This is a challenging problem for humans if their tracks overlap or disappear behind obstacles, and appears to require high-level social intelligence and knowledge to guess what most likely happened when tracks are temporarily hidden.

[12]https://github.com/CMU-Perceptual-Computing-Lab/openpose
[13]https://cmusatyalab.github.io/openface/
[14]https://github.com/opentrack/opentrack

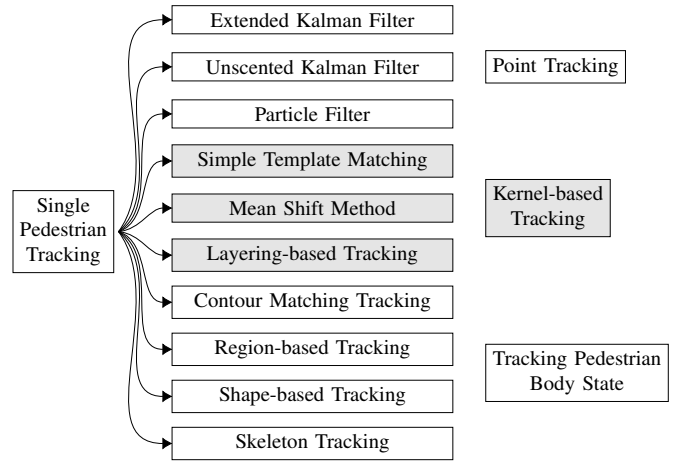

Fig. 7. Single pedestrian tracking models.

Pedestrian tracking consists of two steps: (1) a prediction step to determine several likely next possible pedestrian states, (2) a correction step to check each of these predictions and select the best one. It often requires the estimation of non-linear, non-Gaussian problems due to the nature of human motion, pedestrian sizes, and posture changes [14]. Pedestrian tracking is a challenge for AVs because of the multiple uncertainties (e.g. occlusions) originating from complex environments. Many techniques have been employed for pedestrian tracking, see e.g. [239], [206]. Bar-Shalom et al. [11] presented state estimation algorithms and how they could be applied to tracking and navigation problems. Figure 7 summarizes single pedestrian tracking models.

Previous reviews on tracking methods for pedestrians can be found in [239], [152]. In this section, we first review two classes of methods for single pedestrian location tracking relevant to autonomous vehicle interactions (as previously classified by Yilmaz et al. [239]): point tracking and kernel-based tracking. We then review recent work in the more challenging tasks of body pose tracking and multiple pedestrian tracking. A summary of the tracking methods and papers reviewed in this Part I is provided in the supplementary material Sect. III.

### A. Single Pedestrian Point Tracking

Point tracking typically relies on probabilistic methods based on Bayes filtering [43], [191], [208]. Based on Bayes rule (6), the filter is composed of an initial state, a prediction step and a correction step. The initial state $x_0$ (7) presents the initial belief about the state $x$. The prediction step (8) consists in updating the belief using information about how the target typically moves around. Finally, the correction step (9) updates the state estimate with sensor measurements $z$, to give posterior beliefs $bel(x_t)$ about the state at each discrete time $t$, with a normalizer $\eta$, [186], [219].

$$p(x_t \mid z_t) = \frac{p(z_t \mid x_t)p(x_t)}{p(z_t)} \quad (6)$$

$$bel(x_0) = p(x_0) \quad (7)$$

$$\widehat{bel}(x_t) = \int p(x_t \mid x_{t-1}) \cdot bel(x_{t-1})dx_{t-1} \quad (8)$$

$$bel(x_t) = \eta \cdot p(z_t \mid x_t) \cdot \widehat{bel}(x_t) \quad (9)$$

The transition probability $p(x_t|x_{t-1})$ is of crucial interest as it provides the mathematical bridge from low to high-level pedestrian behavior models. In its lowest form – the standard Kalman filter – it may simply be a Gaussian with zero mean and variance set to model the scale of a (literal) random walk by the pedestrian. But we may have much more predictive information $\theta$ about the pedestrian behavior to form $p(x_t|x_{t-1}, \theta)$. Here $\theta$ could include mid-level information such as the pedestrian's pose, heading, and location on a map. For example, if the pedestrian is standing at the edge of the road, he/she is more likely to wait and cross. Information about the pedestrian's origin and destination could also help to predict the future trajectory. Further information about beliefs, intentions and desires of the pedestrian will also modify the trajectory probability. The transition probability thus provides the interface where all higher-level models, discussed later in Part II [28], will link to low-level pedestrian models. The following are some of the most popular variants of Bayesian Filtering used for pedestrian point tracking:

*a) Kalman Filter (KF):* A KF is a Bayes filter applied to linear systems with continuous states and Gaussian noise $\epsilon_t$,

$$x_t = A_t x_{t-1} + B_t u_t + \epsilon_t, \quad (10)$$

where $A_t$ is the system matrix and $B_t$ is the control matrix.

The measurement probability also depends on a linear model $C_t$ with Gaussian noise $\delta_t$,

$$z_t = C_t x_t + \delta_t \quad (11)$$

where $C_t$ is the measurement matrix.

The prediction step (control update step) increases the uncertainty in the robot's belief, while the measurement update step decreases it.

*b) Extended Kalman Filter (EKF):* An EKF is an extension of the Kalman Filter and approximates non-linear models via Taylor expansion. EKF is a tracking technique well performed in scenarios where there are few changes but it has a computational cost that could be not neglectable for large state and measurement vectors due to the linearization process, which can involve the calculation of big Jacobian matrices. One of the limitations of EKF is that the linearization decreases the accuracy of the system and therefore the pedestrian tracking performance [15]. For example, in [63], the authors try to solve this problem with a CNN detector combined to a Multi-Hypothesis Extended Kalman Filter (MHEKF) for vehicle tracking using low-resolution lidar data.

*c) Unscented Kalman Filter (UKF):* The UKF avoids the linearization problem by a second-order approximation, called the Unscented Transformation. It approximates a probability distribution with chosen weighted points called *sigma points* and estimates its mean and covariance. This leads to better performance in pedestrian tracking, as the Jacobian computation is not necessary anymore, with no or minimum increase of the computational cost [15].

*d) Particle Filter:* This is a sample-based estimator widely used for pedestrian tracking, based on Monte Carlo methods [80], [145], [231]. Unlike EKF, which deals with Gaussian and linearized distributions, it performs state estimation of non-linear and non-Gaussian distributions. It represents the target distribution by a set of samples, called particles. An important step in particle filtering is the resampling, which consists in withdrawing 'weak' particles with low weights from the sample set, and increasing the number of 'strong' particles with high weights [219]. Particle Filtering demands high computation capabilities, when using many particles. A tutorial for implementing particle filters for detection and tracking purposes can be found in [7]. Moreover, Bellotto and Hu [15] evaluated different Bayesian filters, such as EKF, UKF and Sequential Importance Resampling (SIR) particle filter, for people tracking and analysed the trade-off between performance and computational cost of each method.

*B. Single Pedestrian Kernel-based Tracking*

*a) Simple Template Matching:* This is a brute force method. The goal is to compare a region of an image to a reference template image by minimizing the *sum-of-square-difference (SSD)*. For example, in [113], a template matching is proposed for real-time people tracking, which is robust to occlusions and variations of the illumination. In the approach proposed by Lipton *et al.* [137], moving objects are detected in camera images using frame differencing. By combining temporal differencing and template matching, the classified objects are then tracked in real-time on video. In [115], a feature selection method in image sequences is proposed to improve the performance of template matching tracking.

*b) Mean Shift Method:* This is a visual tracking technique trying to match objects in successive frames, where each track is represented by a histogram. The histogram of the region of interest is compared to the histogram of the reference model. The technique iteratively clusters data points to the average of the neighbouring points using a kernel function, similar to $k$-means clustering [44]. In [52], the authors proposed a real-time object tracking using the mean-shift algorithm and the Bhattacharyya coefficient to localize the targets. This method is applied to non-rigid objects tracking observed from a moving camera. Collins [50] applied the mean-shift algorithm to 2D blob tracking and proposed a method to select the kernel scale for an efficient tracking of blobs. In particular, a difference of Gaussian (DOG) mean-shift kernel is chosen to efficiently track blobs through space.

*c) Layering-based Tracking:* Layering consists in splitting an image into several layers by compensating the background motion to estimate the state of a moving object with a 2D parametric model [164]. Each layer is represented by its shape, motion, and appearance (based on intensity) [257]. For instance, in [215], the authors proposed a dynamic layering-based object tracker exploiting spatial and temporal information from its shape, motion and appearance. Their estimation is done using a Maximum-A-Posteriori (MAP) approach with the Expectation Maximisation (EM) algorithm. Layering-based trackers can handle multiple moving objects and occlusion. In [232], a layering-based method is combined with optical flow. A Bayesian framework is used to estimate the layers' appearance and a mixture model is used to segment the image into foreground/background regions. Other layering-based tracking methods applied to imaging sensors can be found in [73], [127].

### C. Body Pose State Tracking

Tracking the whole state of a pedestrian's body – including skeleton pose, head direction, feet and walking directions – may provide useful information about the pedestrian's state and intention. These silhouette tracking methods are based on an accurate shape description of the pedestrian object. The general technique consists in finding the pedestrian region in each frame with an object model computed from the previous frames. The advantage is that it can cope with different types of shape, occlusion problems, etc.

*a) Contour Matching Tracking:* Tracking is performed considering the contours of objects, which are dynamically updated in successive frames. Geiger *et al.* [93] proposed a contour tracking method that is based on Dynamic Programming (DP) to detect and track the contour of multiple shapes and provide the optimal solution to the problem. Techmer [217] developed a real-time approach to contour tracking relying on the distance transformation of contour images and tested it on real-world images. Baumberg and Hogg [13] proposed a method that combines dynamic filtering (Kalman filter) with an active shape model to track a walking pedestrian in real-time. However, this tracking technique is very sensitive to the initialization, so other solutions have been developed to overcome that issue [240].

*b) Region-based Tracking:* This technique is based on the color distribution of objects. In [1], a tracking algorithm is proposed based on multiple fragments of object images, creating a histogram of the current frame that is compared to the histogram of the patches. Their method is able to handle occlusion and pose changes in an efficient manner. Other methods have employed depth, probabilistic occupancy maps and gait features to estimate a region's features, but in some cases (e.g. depth features) this requires the computation of multiple views of the same scene. Meyer and Bouthemy [146] developed a method to track objects over a sequence of images using a recursive algorithm based on image regions information, such as their position, shape and motion model.

*c) Shape Matching Tracking:* Shape matching tries to match silhouettes found in two consecutive frames. Performed with Hough transform, it can handle occlusion problems. For instance, in [51], a silhouette-based model is used to identify people from their body shape and gait.

*d) Skeleton Tracking for Body Language and Gesture Recognition:* Skeleton tracking, based on tracking human body parts, is a popular technique [92], [238], [196], [153]. Schwarz *et al.* [199] presented a full-body tracker using depth data from a Kinect sensor. 3D data is represented by a graph structure which can deal with variations in pose and illumination. A skeleton is then fitted to the 3D data by constrained inverse kinematics and geodesic distances between body parts. Sinthanayothin *et al.* [205] reviewed skeleton tracking methods using Kinect sensors. Make Human Community[15] is an open-source project building parametric models of humans based on realistic skeleton structures, mainly targeted at video games users, but also used as a generative machine vision and tracking model for 3D sensor data.

### D. Multiple Pedestrian Tracking

Multiple pedestrian tracking (a form of MTT, Multi-Target Tracking) names the task of (rather than specific algorithm for) tracking the poses of several pedestrians at the same time. The pedestrians may be close, overlapping, or obstructing one another, and they may be indistinguishable from one another other than by their pose. This is required for AV interactions with multiple pedestrians, ranging from two well-separated pedestrians, to small groups of pedestrians (often crossing roads together) and to dense crowds. MTT creates a data association problem: how to know which pedestrian detection belongs to which track? A probabilistic MTT model would maintain beliefs at each time step about the state of every track and consider every possible association of detections to tracks; then, it would perform inference accordingly. However, the number of associations grows exponentially with the number of pedestrians, so this approach is unlikely to work in very crowded scenarios. Standard approximations then include making hard 'winner-take-all' assignments at each time step; maintaining search trees of recent possible assignments; and pruning association hypotheses. There are many possible variations on these approximations, all making use of basic individual-pedestrian trackers as components.

Leal-Taixé *et al.* [129] presented a benchmark for Multiple Object Tracking that was launched in 2014 and callled *MOTchallenge*. This benchmark provides a framework for evaluating the performance of state-of-the-art MTT algorithms. About 50 methods have been tested up to now on this benchmark. However, [129] does not describe these algorithms, while Fan *et al.* [78] only presents a survey on visual methods. A previous review on multiple object tracking was proposed in [142]. The remainder of this section will therefore extends their work for multiple person tracking and try to give an overview of the main methods, challenges and future directions of MTT techniques, which intelligent transportation systems heavily rely upon. Figure 8 summarizes the techniques described in this section.

*1) Categories of MTT methods:* The following paragraphs will develop the different categories of multi-target tracking methods that are defined according to their initialization method used, the processing method, or the tracking output.

---

[15]http://www.makehumancommunity.org/

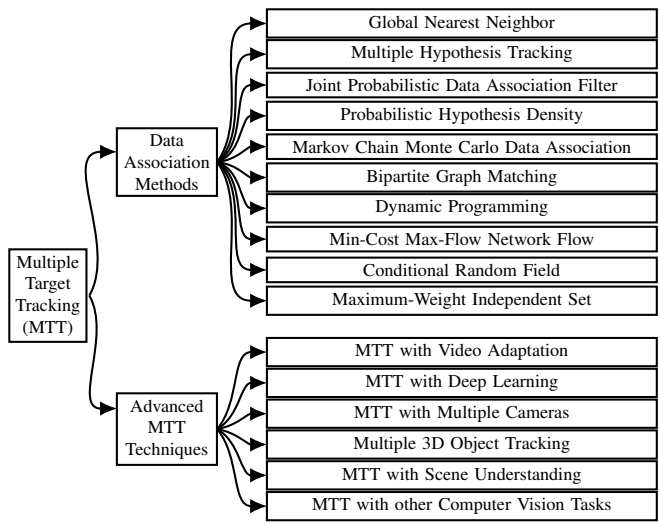

Fig. 8. MTT data association and advanced techniques.

*a) Initialization Method:* The first category is characterised by the detection technique used before tracking. The most commonly used method is Detection-Based Tracking (DBT) where a program is trained in advance to detect the target object in the input data (e.g. images) [111]. This technique can deal with a variable number of target objects, but it cannot track unknown objects that were not part of the training. The other initialization method is Detection-Free Tracking (DFT), which requires manual initialization, i.e., an operator labels manually the target objects. In this case, the object detection is error-free but the tracking can usually only deal with a fixed number of target objects. Neiswanger *et al.* [157] proposed a method to track multiple people in video sequences without any pre-defined person detector. A Dirichlet process is used to find the clusters in the images and then a Sequential Monte Carlo (SMC) method with local Gibbs iterations and a Particle Markov Chain Monte Carlo (PMCMC) are used to infer the posterior of targets. Lin *et al.* [135] developed a detection-free multiple target tracking method which relies on video bundle representation and a spatio-temporal graphical model to infer the trajectories of people.

*b) Processing Model:* This second category refers to the information processing mode: online or offline tracking. Online tracking [120] is a sequential tracking, which relies on up-to-date information. It is a causal method where only past and current observations are used. Offline tracking [118] instead uses information both from past and future observations, therefore it is not causal. In order to estimate the output, offline tracking needs to evaluate all the observations from all the frames, which requires a high computation cost. The manual assignment guarantees a tracking process free of false detections, but is not suitable for real-time applications. Both online and offline tracking methods are proposed in [242].

*c) Tracking Output:* MTT methods can be grouped according to output. Output results are fixed for MTT methods relying on deterministic optimization, i.e., there is no randomness when these methods are run many different times, whereas for probabilistic optimization methods, output may vary for several trials cf. section V-D3.

*2) Challenges of MTT Approaches:* There are multiple challenges with the tracking of multiple objects. Here we summarise the most important ones.

*a) Similarity Measurement:* The first problem is how to measure the similarity between objects in different frames. Different models have been proposed to deal with the *similarity measurement* between objects. The most commonly-used technique in visual tracking relies on the object's appearance, i.e., its visual features. There are local features, which can be obtained by the KLT algorithm or optical flow (if we treat each pixel as the finest local range) to get information about object motion patterns [202]. Region features are extracted from an image and represented by a bounding box. Three main types of region features exist: zero-order, first-order and up-to-second-order type. The zero-order type represents region features as color histogram or raw pixel templates. Although color is a common similarity measure, the problem is that it does not take into account the spatial layout of the object region. A first-order type uses gradient-based representations or level-set formulation to deal with region features [47]. Gradient-based representation is a robust technique because it describes well the shape of the object and it is less dependent to illumination conditions, but it cannot handle occlusion problems. An up-to-second-order type computes region covariance matrices to model the observed features [172]. This is a robust strategy but it requires a high computation capability.

*b) Track Identification:* The second problem consists in recovering the identity of objects from the similarity measurement across frames. Different strategies *compute the similarity* between objects. A survey on similarity measures for probability density functions is provided in [250]. In case of a single cue, a distance measure is computed from two color histograms and then transformed into similarity using the exponential function or an affinity measure such as the Normalized Cross Correlation (NCC). When multiple cues are available, there are several strategies used to fuse the information [26]. *Boosting*, for example, consists in selecting the most representative features from a large set of proposed features using a machine learning algorithm such as AdaBoost [236]. *Concatenation* uses features from different cues and concatenates them for computation. *Summation* takes affinity values from different features and adds a weight to each value. *Product strategy* assumes independence between affinity values and computes their weighted product. *Cascading* uses diverse visual representations and tries to determine the finest model appearance [224]. To improve tracking prediction, exclusion models can be used to prevent physical collisions, assuming that two distinct pedestrians cannot be at the same place at the same time. Two types of constraints can be applied to the trajectory hypotheses: detection-level exclusion and trajectory-level exclusion [142]. Detection-level exclusion assumes that two detections in a frame cannot be assigned to the same target. Trajectory-level exclusion means that two trajectories cannot be too close to each other. In order to avoid that, a penalty is assigned to two hypotheses that are too close and which have different trajectories, to suppress one of them.

*c) Occlusion:* The third problem is how to handle *occlusions* of tracking targets. Three major strategies are employed to face this challenge. *Part-to-whole* divides the object into several parts and then computes an affinity for each part. When an occlusion occurs, only the unoccluded parts are taken into account for estimation [210], [237]. In *hypothesize-and-test*, detection hypotheses are generated for two objects with different levels of occlusion, which are then tested for example using MAP or a multi-person detector [213]. The *buffer-and-recover* technique keeps the states of objects over several frames, before and during an occlusion. When it ends, the states of objects are recovered using the observations on the frame buffer [189].

*3) Multi-Tracks and Data Association Methods:* Probabilistic or deterministic optimization are the common methods to deal with multiple tracks and data association problems. Data association is about the uncertainty related to measurements, it aims at associating observed measurements with current known tracks or generate new tracks. Deterministic optimization methods are usually suitable for offline tracking, as they require observations from several or all the frames in advance [142], whereas probabilistic methods are commonly used for online or real-time tracking. Bar-Shalom and Li [10] presented several data association algorithms, such as Nearest Neighbors (NN), Multi-Hypothesis Tracking (MHT), Joint Probabilistic Data Association Filter (JPDAF), or Probability Hypothesis Density (PHD), and evaluated their performances.

*a) Global Nearest Neighbour (GNN):* GNN [22] is one of the simplest methods for data association. At every new time step, it 'hardly' assigns each current observation to a single best object without revising the past. In [124], GNN is described as a 5-step algorithm: (1) receive data for each scan; (2) each track is first defined as a cluster and if common observations are found for two tracks, they are merged into a 'super cluster'; (3) observations are assigned to each cluster using Munkres algorithm [126]; (4) tracks' states are updated using some estimation technique such as Kalman filter; (5) observations which are not associated to any existing tracks are used to create new tracks. The work in [8] developed a multiple person tracker where GNN is used for data association with a new distance function and a Kalman filter for state estimation. The proposed method is suitable for occlusion issues.

*b) Multiple Hypothesis Tracking (MHT):* This filter, originally proposed by Reid [180], is an iterative algorithm which can handle multiple tracking targets, with occlusions, and give optimal solutions. It makes predictions on each hypothesis for the succeeding frame. Each hypothesis represented a group of mutually separate tracks [219]. The aim of MHT is to overcome the wrong data association problem by representing the posterior belief with a mixture of Gaussians, where each Gaussian component is considered to be a track and relies on a unique data association decision. MHT is a more complex approach than GNN: it propagates assignment probabilities over time as a tree of the future observations in order to resolve past ambiguities. Luber *et al.* [141] proposed a model that uses social force model as a motion model for MHT. Motivations, principles and implementations of MHT are presented in [21]. MHT is generally considered to be too slow and memory-expensive for multi-target tracking methods as pruning and priming have to be applied in order to keep the size of the tree manageable [121]. Amditis *et al.* [4] proposed examples of MHT implementation for MTT using laser scanner data.

*c) Joint Probabilistic Data Association Filter (JPDAF):* This method has been proposed by [87]. It generates multiple tracks-to-measurement hypotheses and calculates the hypotheses probabilities. Then, it gives hard, unrevisable assignment of hypotheses that are merged to each track at each time step. This is more complex than GNN because the latter is greedy and just assigns each observation individually to its nearest object, while JPDAF allows some entanglement over space. In contrast, MHT filter allows some entanglement over time [4], considering all the joint data-object assignments and picking the best. JPDAF runs faster than MHT [255], but it requires a fixed number of targets. Chen *et al.* [41] proposed the use of a JPDAF to compute hidden Markov models transition probabilities for a contour-based human tracking method performing in real-time. Liu *et al.* [139] proposed a person tracking method combining JPDAF and multi-sensor fusion. [106] implemented a tracking method based on JPDAF and capable of tracking about 400 persons in real-time. Rezatofighi *et al.* [183] presented a JPDAF-based tracker for challenging conditions, such as observations from fluorescence microscopy sequences or surveillance cameras.

*d) Probabilistic Hypothesis Density (PHD):* This filter was introduced by [49]. It can track a variable number of tracks, estimating their number and their locations at each time step. There are different types of PHD filters, such as the Sequential Monte Carlo PHD filter (SMC-PHD) [187], the Gaussian Mixture PHD filter (GM-PHD) [244] and the Gaussian Inverse Wishart PHD filter (GIW-PHD) [99]. Zhang *et al.* [246] used a GMM-PHD (Gaussian Mixture Measurement PHD) tracker to tackle problems with bearing measurements. Khazaei *et al.* [119] developed a PHD filter in distributed camera network where each camera fuses its track estimates with its neighbors. Feng *et al.* [81] proposed a variational Bayesian PHD filter with deep learning update to track multiple persons. In [57], a PHD filter is used to track in real-time multiple people in a crowded environment. Yoon *et al.* [241] used hybrid (i.e. local and global) observations in a PHD filter, where the filter observations are combined with local observations generated by on-line trained detectors. This method allows to handle missed detections and it assigns an identity to each person.

*e) Markov Chain Monte Carlo Data Association (MCM-CDA):* Introduced first by [168], this filter is an approximation of the Bayesian filter, derived from MCMC, which draws a set of samples and builds Markov chains over the target state space. A sampler moves from its current state to the next following the proposal distribution. The new state is accepted with an acceptance probability, otherwise the sampler stays at its current state. Oh *et al.* [159], [158] proposed an MCMCDA algorithm known as Metropolis-Hastings, where single-scan and multi-scan MCMCDA algorithms are used for known and unknown number of targets, respectively. A bipartite graph is used to represent possible associations between observations and targets. Their simulation results show a better performance than MHT algorithms and their method has been tested on tracking people from video sequences. Yu *et al.* [243] proposed a data-driven MCMC (DD-MCMC) approach for sampling and incorporating a person's motion and appearance information, using a joint probability model. Their method was tested in simulations and on real videos.

*f) Bipartite Graph Matching:* This uses two sets of graph nodes representing existing trajectories and new detections in online tracking, or two sets of tracklets (components of tracks) in offline tracking. The weights of nodes model affinities between trajectories and detections. The Bipartite assignment algorithm or optimal Hungarian algorithm is used to find matching nodes in the two sets. A review on graph matching is presented in [53]. Chen *et al.* [109] used a dynamical graph matching method to track multiple people in order to dynamically change the graph nodes with the tracks movements.

*g) Dynamic Programming:* This method solves the data association problem by linking several detections over time. Pirsiavash *et al.* [170] used a greedy algorithm based on dynamic programming to find the global solution in a network flow. Another method is presented in [18] which can follow up to six people over several frames.

*h) Min-Cost Max-Flow Network Flow:* This is a popular method, which models the network flow as a directed graph. A trajectory is represented by a start node and an end node (sink), and it corresponds to one flow path in the graph. The global optimal solution is obtained with the push-relabel algorithm. Zhang *et al.* [245] used a min-cost flow algorithm combined with a recursive occlusion model to deal with occluded people. Their method does not require pruning. Chari *et al.* [38] proposed a new approach to the min-cost max-flow network flow optimization using pair-wise costs, which can deal with occluded people.

*i) Conditional Random Field (CRF):* A graph $G = (V, E)$ is defined as a set of nodes $V$ and a set of edges $E$. Nodes represent observations and tracklets. A label is used to predict which track observations are linked to. Sutton and McCallum [211] presented a CRF tutorial. Taycher *et al.* [216] proposed a person tracking method learning from data, based on a CRF state-space estimation and a grid-filter with real-time capabilities. Milan *et al.* [147] developed a CRF-based MTT, detecting people using a HOG-SVM detector, and defining two unary potentials for detection and superpixel nodes. Milan *et al.* [149] proposed a CRF-based multiple person tracker using discrete-continuous energy minimization, whose goal is to assign a unique trajectory to each detection.

*j) Maximum-Weight Independent Set (MWIS):* The MWIS graph is defined as $G = (V, E, w)$. As in the CRF, the nodes $V$ represent the pairs of tracklets in successive frames, which are given a weight $w$ indicating the affinity of the tracklet pair. If two tracklets share the same detection, then their edges $E$ are connected together. Brendel *et al.* [25] proposed a multi-target tracker based on MWIS data association algorithm. Their approach is as follows: (1) detection of multiple targets in all frames using different object detectors; (2) detections are considered as distinct tracks, with the assumption that one detection can only be one track; (3) a graph is built to match tracks over two consecutive frames; (4) an MWIS algorithm is used to perform the data association with guaranteed optimal solution; (5) statistical and contextual properties of objects are learnt online for their similarity measurement using Mahalanobis distances; steps (2) to (5) are repeated over the frames to handle long-term occlusions by merging or splitting tracks. In [105], a multi-person tracker is used with data association modelled as a Connected Component Model (CCM) based on MWIS. A divide-and-conquer strategy is used to solve the Multi-Dimensional Assignment (MDA) problem.

*4) Advanced MTT Techniques:* Here *Advanced MTT* refers to multi-target tracking that is performed at a higher-level, simultaneously with other tasks.

*a) MTT with Video Adaptation:* MTT approaches rely on an object detector that is trained offline, so its performance can be totally different from a video to another. A possible solution is to create a generic detector adapted for a specific video by tuning some parameters. Previous works for multiple people tracking include [91], [39].

*b) MTT with Deep Learning:* Deep learning has proven to be a high performance method for classification, detection and many computer visions tasks. Applied to MTT, deep learning could provide a stronger observation model which could increase the tracking accuracy [242], [131]. In [161], Ondruska *et al.* introduced deep tracking, an end-to-end human tracking approach, based on recurrent neural network, using unsupervised learning on simulated data without dealing with the data association problem. In [66], Dequaire *et al.* used a similar method for static and dynamic person tracking in real-world environments. In [148], Milan *et al.* proposed a complete online multiple people tracking method based on recurrent neural networks.

*c) MTT under Multiple Cameras:* Also called Multi-Target Multi-Camera (MTMC), this type of systems can be used to improve large tracking problems. Wang *et al.* [228] presented a survey on the challenges of MTMC. One problem would be overlapping cameras, in which case it is necessary to find a good way to fuse multiple information. But if the camera angles do not overlap, then the data association problem becomes an identification problem. In [184], Ristani *et al.* proposed different performance measures to test MTMC methods. In [185], they used neural networks to learn features from MTMC systems and for re-identification. In [140], Generalized Maximum Multi-Clique optimization – a graph-based method – is used for the MTMC problem. Munaro *et al.* [155] developed an open-source software, called OpenPTrack, for multi-camera calibration and people tracking using RGB-D data.

*d) Multiple 3D Object Tracking:* This method could provide better position accuracy, size estimation and occlusion handling. The major problem for this technique is the camera calibration. Park *et al.* [167] applied 3D object tracking from a monocular camera for augmented reality applications. Some other works on 3D visual tracking include [59], [166], [194], which used a single camera with a multi-Bernoulli mixture tracking filter. Some works with 3D lidar sensors include [114], [207], [234], which proposed online classification of humans for 3D lidar tracking. In [175], both camera and lidar data are used to improve people tracking.

*e) MTT with Scene Understanding:* Scene understanding can provide contextual information and scene structure for the tracking algorithm, especially in crowded scenes. Leal-Taixé et al. [128] developed a model that decomposes an image and extracts features from the observed scene called 'interaction feature strings'. These features are then used in a Random Forest framework to track human targets [64].

*f) MTT with Other Computer Vision Tasks:* Information from image segmentation or human pose estimation could not only improve the performance of multiple-people tracking but also the computation of the tracking algorithm. For example, in [147], tracking is done with image segmentation and in [47] people are tracked for group activity recognition.

*E. Discussions*

Single pedestrian tracking is now a fully mature area with widely available open-source and commercial implementations. Body pose tracking has made strong recent progress, likely to soon bring it to maturity, through the use of larger data sets and computer power.

Tracking multiple pedestrians requires additional algorithms which were major research areas until recently, but have largely matured in the last few years with methods such as MHT becoming standard. Tracking multiple pedestrians in the presence of occlusion by one another or by other objects remains a serious research problem, which requires the use of other data or prior information to compensate for the lack of purely visual data. We suggest that the higher-level models from psychology and sociology discussed in the Part II of this review [28] should be used to provide such priors. Traditionally, tracking was a clearly separate task from both lower (detection) and higher (behaviour modelling) layers of pedestrian modelling, but a current trend is to merge it with nearby layers through neural network and probabilistic methods in this fashion to improve performance.

Practical implementation of tracking algorithms may be found in the Bayes Tracking library[16] which provides open-source implementation of EKF, UKF and SIR Particle Filters with NN and JPDA data association algorithms. In addition, a detection and tracking pipeline[17] contains an implementation of MHT. Choi et al. [45] proposed a fast tracker TRACA[18] with a deep feature compression approach for single target tracking.

In terms of computational efficiency, Bellotto and Hu [15] have shown that Kalman-based people tracking is much faster than particle-based, and in particular that UKF was faster and still almost as reliable as particle filter. Linder et al. [136] proposed a comparison (computation speed and other metrics) of various people tracking methods, including NN trackers, MHT and others. A common heuristic for some mobile robots is to run at 10Hz or more, i.e. if the robot moves at 1m/s, a people tracker running at 10Hz will estimate the position of humans every 10cm, which is usually considered safe. But with cars moving much faster such as 10m/s (36km/h), the computational requirements would be greater, such as operating 100Hz to obtain the same 10cm accuracy.

## VI. CONCLUSIONS

Autonomous vehicles must interact with pedestrians in order to drive safely and to make progress. It is not enough to simply stop whenever a pedestrian is in the way as this leads to the freezing robot problem and to the vehicle making no progress. Rather, AVs must develop similar interaction methods as used by human drivers, which include understanding the behaviour and predicting the future behaviour of pedestrians, predicting how pedestrians will react to the AVs movements, and choosing those motions to efficiently control the interaction.

This Part I review has surveyed the state of the art in the lower levels of machine perception and intelligence needed to enable such interaction control, namely: sensing, detection, recognition, and tracking of pedestrians. It has found that the level of maturity of these fields is high at the lowest levels, but fades into current research areas at the higher-levels. Sensing technology has progressed to maturity over the last decade so that lidars and stereo cameras are now reliable and cheap enough for use in research and even by hobbyist systems. Similarly, GPUs have fallen in price to enable both stereo camera processing and deep learning recognition to be run in these systems. Deep learning recognition has largely replaced classical feature-based methods for detection. Open-source software is mature and freely available for these tasks.

Beyond detection are areas with successful, open-source, partial implementations but which require further research to become fully mature. Recognition of body pose and head direction are almost mature, including via deep learning methods. But recognition of higher-level states, such as gestures used for explicit signalling, body language used as implicit signalling, actions as sequences of poses, and recognition of underlying emotional state, remain research areas.

Tracking is mature for single pedestrians, but remains challenging for multiple pedestrians in the presence of occlusion. Algorithms to solve this task are known but require the use of extensive prior knowledge to predict behaviour in the absence of sensory information, which is not yet fully available. This includes information from recognition of poses, gestures, actions, and emotions, but also feedback information from very high-level models of behaviour and psychology which will be studied in Part II of this review [28].

---

[16] https://github.com/LCAS/bayestracking
[17] https://github.com/sbreuers/detta
[18] https://github.com/jongwon20000/TRACA

# Supplementary Material: Pedestrian Models for Autonomous Driving Part I: Low-Level Models, from Sensing to Tracking

Fanta Camara[1,2], Nicola Bellotto[2], Serhan Cosar[3], Dimitris Nathanael[4], Matthias Althoff[5], Jingyuan Wu[6], Johannes Ruenz[6], André Dietrich[7] and Charles W. Fox[1,2,8]

## I. QUALITY OF CITATIONS

These linked papers (Part I and II) review over 450 papers from high quality journals and conferences such as *CVPR, ICRA, PAMI, IROS, ITSC, ECCV, IV*. It is common in Computer Science fields including machine vision and machine learning for conferences to be considered higher quality or similar quality to journals, while psychology and sociology fields typically consider journals to be more authoritative. The following figures give some ideas about the quality of the cited papers.

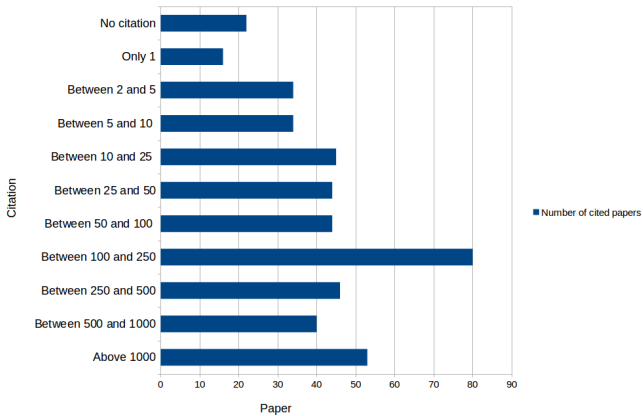

Fig. 1. Number of citations per paper

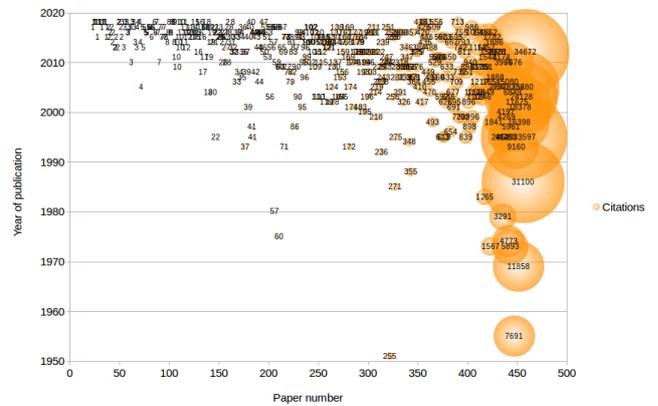

Fig. 2. Number of citations per paper and per year of publication

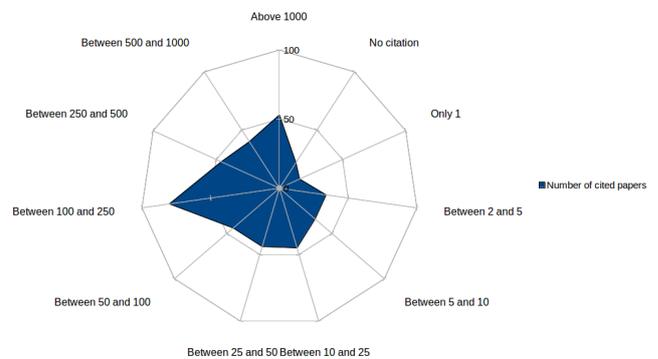

Fig. 3. Number of citations per paper

This project has received funding from EU H2020 interACT (723395).
[1] Institute for Transport Studies (ITS), University of Leeds, UK
[2] Lincoln Centre for Autonomous Systems, University of Lincoln, UK
[3] Institute of Engineering Sciences, De Montfort University, UK
[4] School of Mechanical Engineering, Nat. Tech. University of Athens
[5] Department of Computer Science, Technische Universität München
[6] Robert Bosch GmbH, Germany
[7] Chair of Ergonomics, Technische Universität München (TUM), Germany
[8] Ibex Automation Ltd., UK
Manuscript received: 2019-03-11; Revisions:2019-10-21, 2020-02-25.
Accepted: 26-06-2020.

## II. SUMMARY OF PEDESTRIAN RECOGNITION MODELS

TABLE I: Summary of the recognition models

| Study/Paper | Input/Evaluation | Method | Recognition Models | Additional Info | SAE Level |
|---|---|---|---|---|---|
| Cao et al. [1] [2] | Images | CNN model with Part Affinity Fields (PAF) | Pose estimation | OpenPose: open-source software | Level 2 |
| Shotton et al. [3] | Motion capture and synthetic data | Body parts representation model | 3D human pose estimation | | Level 2 |
| Iqbal et al. [4] | Video data | Graphical model | Pose estimation and tracking | Release of PoseTrack a new dataset | Level [2,3] |
| Tompson et al. [5] | Monocular images | Deep CNN model with Markov Random Field | Pose estimation | | Level 2 |
| Fragkiadaki et al. [6] | Motion capture data: H3.6M dataset [Ionescu et al. 2014] | Encoder-Recurrent-Decoder (ERD) | Body pose estimation | | Level [2] |
| Martinez et al. [7] | Motion capture data: H3.6M dataset [Ionescu et al. 2014] | Recurrent neural network with a gated recurrent unit (GRU) | Body pose estimation | | Level 2 |
| Tang et al. [8] | Motion capture data: H3.6M dataset [Ionescu et al. 2014] | Deep neural network (modified High-way Unit (MHU)) | Pose estimation | | Level 2 |
| Ghosh et al. [9] | Motion capture data: datasets in [Ionescu et al. 2014] and [Holden et al. 2016] | Dropout AutoEncoder LSTM (DAE-LSTM) | Pose estimation | | Level 2 |
| Ma et al. [10] | Images | CNN model | Body heading | No annotations needed | Level 2 |
| Kohari et al. [11] | Video | CNN model | Body orientation | Service robot | Level 2 |
| Darrell et al. [12] | Images | Statistical model | Head direction | From a mobile robot | Level [2,3] |
| Schulz et al. [13] | Grayscale images | Multi-classifiers | Head pose | | Level 2 |
| Benfold et al. [14] | Video | HOG and colour features | Gaze tracking | | Level 2 |
| Baltrusaitis et al. [15] | Video | Deep learning model | Head pose and eye-gaze estimation | Openface: open source software | Level 2 |
| Cornejo et al. [16] [17] | Images | Principal Component Analysis + Gabor wavelets or CENTRIST features | Emotion recognition | | Level 2 |
| Cambria et al. [18] [19] | Images | Review paper | Sentiment analysis | | [Level 2,3] |
| Poria et al. [20] | Videos | CNN model with recurrent multilayer kernel learning | Emotion recognition | | Level 2 |
| Horng et al. [21] | Video | Dynamic template matching | Driver fatigue detection | | Level 2 |
| Denuyl et al. [22] | Video | | Face expression recognition | FaceReader: commercial product | Level 2 |
| Ahmed et al. [23] | Images | Deep neural networks | Re-identification | | Level 2 |
| Zheng et al. [24] | Images | Bag of Words model | Re-identification | | Level 2 |
| Zheng et al. [25] | Images | CNN model | Re-identification | Unlabeled images | Level 2 |
| Li et al. [26] | Images | Filter airing neural network (FPNN) model | Re-identification | Occlusion handling | Level 2 |
| Chen et al. [27] [28] [29] | Images | Hidden Markov model (HMM) | Gesture recognition | | Level 2 |
| Freeman et al. [30] | Images | Orientation histograms | Gesture recognition | 10 different hand gestures recognition | Level 2 |
| Ren et al. [31] | Images | Template matching with Finger-earth Mover's Distance (FEMD) | Gesture recognition | | Level 2 |
| Quintero et al. [32] | Images | Hidden Markov models (HMM) | Body language Recognition | | Level [2,3] |
| Wang et al. [33] | Images | Background subtraction + PCA | Body language Recognition | | Level 2 |
| Chaaraoui et al. [34] | Videos | Contour points | Activity recognition | Real-time method | Level 2 |
| Dollár et al. [35] | | Spatio-temporal features | Activity recognition | | Level 2 |
| Vail et al. [36] | Videos | Hidden Markov models and Conditional random field | Activity recognition | | Level 2 |
| Liu et al. [37] | RGB data | Coupled conditional random field | Activity recognition | | Level 2 |
| Coppola et al. [38] | RGB-D data | Dynamic Bayesian mixture model (DBMM) | Activity recognition | | Level 2 |

## III. SUMMARY PEDESTRIAN TRACKING MODELS

TABLE II: Summary of pedestrian tracking models

| Study/Paper | Input/Evaluation | Method | Tracking Models | Additional Info | SAE Level |
|---|---|---|---|---|---|
| Del Pino et al. [39] | Low resolution LiDAR data | Multi-Hypothesis EKF (MHEKF) | Point Tracking | | Level 2 |
| Bellotto et al. [40] | Robot with laser and camera | EKF, UKF, SIR Particle filter | Point Tracking | Trade-off between performance and computation cost | Level 2 |
| Arulampalam et al. [41] | Example | Particle filter implementations | Point Tracking | | Level 2 |
| Fen et al. [42] | Video data | Color histogram based particle filter | Point Tracking | | Level 2 |
| Jurie et al. [43] | Video data | Template matching with SSD | Kernel-based Human Tracking | Real-time method and robustness to occlusions and illuminations | Level 2 |

IEEE TRANSACTIONS ON INTELLIGENT TRANSPORTATION SYSTEMS                                                                                   3TABLE II: Summary of pedestrian tracking models

| Study/Paper | Input/Evaluation | Method | Tracking Models | Additional Info | SAE Level |
|---|---|---|---|---|---|
| Lipton et al. [44] | Video data | Frame differencing + Template matching | Kernel-based Human Tracking | Real-time method | Level 2 |
| Kaneko et al. [45] | Image sequences | Template matching with a feature selection method | Kernel-based Human Tracking | | Level 2 |
| Comaniciu et al. [46] | Moving camera data | Mean-shift algorithm with Bhattacharyya coefficient | Kernel-based Human Tracking | | Level 2 |
| Collins et al. [47] | Video data | Mean-shift algorithm with 2d blob tracking | Kernel-based Human Tracking | | Level 2 |
| Tao et al. [48] | Airborne vehicle tracking system | Dynamic layering method + MAP using EM algorithm | Kernel-based Human Tracking | | Level 2 |
| Yalcin et al. [49] | Image sequences | Layering method with optical flow | Kernel-based Human Tracking | | Level [2,3] |
| Geiger et al. [50] | Image sequences | Contour matching method based on Dynamic programming | Tracking pedestrian body state | | Level 2 |
| Techmer et al. [51] | Real-world images | Contour tracking with distance transformations of contour images | Tracking pedestrian body state | | Level 2 |
| Baumberg [52] Yilmaz [?] | Image sequences | Dynamic Kalman filter with active shape model | Tracking pedestrian body state | Method sensitive to initialization | Level 2 |
| Adam et al. [53] | Image sequences | Region color histogram method | Tracking pedestrian body state | Occlusion and pose changes handling | Level 2 |
| Meyer et al. [54] | Image sequences | Recursive algorithm using image regions information | Tracking pedestrian body state | | Level 2 |
| Collins et al. [55] | Gait databases: CMU, MIT, UMD, USH | Silhouette based model | Tracking pedestrian body state | To identify people from their body and gait | Level 2 |
| Schwarz et al. [56] | Kinect data | Graph method with skeleton fitting | Tracking pedestrian body state | Full-body tracker | Level 2 |
| Sinthanayothin et al. [57] | Kinect data | Skeleton tracking | Tracking pedestrian body state | Review paper | Level 2 |
| Konstantinova et al. [58] | 5 test matrices | Global Nearest Neighbor with Munkres algorithm | Multi-Target Tracking (MTT) | | Level [2, 3, 4] |
| Azari et al. [59] | IBM, PETS2000 and PETS2001 databases | Kalman filter with Global Nearest Neighbor (GNN) | Multi-Target Tracking (MTT) | Occlusion handling | Level [2,3,4] |
| Reid et al. [60] | Monte Carlo simulation | Iterative algorithm with Multi-Hypothesis Tracking(MHT) | Multi-Target Tracking (MTT) | Occlusion handling | Level [2,3] |
| Luber et al. [61] | Two datasets collected in indoor and outdoor environments | Social force with Multiple Hypothesis Tracking (MHT) | Multi-Target Tracking (MTT) | | Level [3,4] |
| Kim et al. [62] | PETS and MOTChallenge benchmarks | Multiple Hypothesis Tracking (MHT) | Multi-Target Tracking (MTT) | | Level [2,3] |
| Zhou et al. [63] | Computer simulations | Joint probabilistic data association filter (JPDAF) with a depth-search approach | Multi-Target Tracking (MTT) | | Level [2,3] |
| Chen et al. [64] | Video data | Contour based tracker with JPDAF and HMM | Multi-Target Tracking (MTT) | Real-time method | Level [2,3] |
| Liu et al. [65] | Simulations and real robot | Sample-based JPDAF and multi-sensor fusion | Multi-Target Tracking (MTT) | Real-time method | Level [2,3] |
| Horridge et al. [66] | Simulations | JPDAF based tracker | Multi-Target Tracking (MTT) | 400 tracks in real-time | Level [2,3,4,5] |
| Rezatofighi et al. [67] | Fluorescence microscopy sequences and surveillance camera data | JPDAF based tracker | Multi-Target Tracking (MTT) | | Level [2,3] |
| Zhang et al. [68] | Simulations | Gaussian Mixture Measurement PHD tracker (GMM-PHD) | Multi-Target Tracking (MTT) | Handle bearing measurements | Level [2,3] |
| Khazaei et al. [69] | Data from a distributed network of cameras | Probabilistic Hypothesis Density (PHD) filter based tracker | Multi-Target Tracking (MTT) | | Level [2,3] |
| Feng et al. [70] | Simulations with sequences from CAVIAR dataset | Variational Bayesian PHD filter with deep learning updates | Multi-Target Tracking (MTT) | | Level [2,3] |
| Correa et al. [17] | Tested on a real-time crowded environment | PHD filter | Multi-Target Tracking (MTT) | | Level [4, 5] |
| Yoon et al. [71] | ETH dataset | Sequential Monte Carlo PHD filter (SMC-PHD) | Multi-Target Tracking (MTT) | Can handle missing detections | Level [2,3] |
| Oh et al. [72] [73] | Simulations | Markov Chain Monte Carlo Data Association (MCMCDA) Metropolis-Hastings method | Multi-Target Tracking (MTT) | | Level [2,3] |
| Yu et al. [74] | Simulations and video data | Data-driven Markov Chain Monte Carlo data association (DD-MCMCDA) | MTMulti-Target Tracking (MTT) | | Level [2,3] |
| Chen et al. [75] | Video data | Dynamical graph matching | Multi-Target Tracking (MTT) | Tracker can deal with interactions | Level [2,3] |
| Pirsiavash et al. [76] | Video data | Greedy algorithm based on Dynamic Programming | Multi-Target Tracking (MTT) | | Level [2,3] |
| Zhang et al. [77] | CAVIAR and ETHMS datasets | Min-Cost Flow algorithm with an explicit occlusion model (EOM) | Multi-Target Tracking (MTT) | Occlusion handling | Level [2,3] |

TABLE II: Summary of pedestrian tracking models

| Study/Paper | Input/Evaluation | Method | Tracking Models | Additional Info | SAE Level |
| --- | --- | --- | --- | --- | --- |
| Chari et al. [78] | PETS and TUD datasets | Min-Cost Max-Flow network optimization with pair-wise costs | Multi-Target Tracking (MTT) | Occlusion handling | Level [2,3] |
| Taycher et al. [79] | Video data | Conditional Random Field (CRF) state estimation and grid filter | Multi-Target Tracking (MTT) | Real-time capability | Level [2,3] |
| Milan et al. [80] | PETS 2010 Benchmark and TUD-Stadtmitte dataset | CRF-based multiple tracker with HOG-SVM detector | Multi-Target Tracking (MTT) | | Level [2,3] |
| Milan et al. [81] | PETS, TUD, ETHMS datasets | CRF-based multi-target tracker using discrete continuous minimization | Multi-Target Tracking (MTT) | Trajectory estimation of targets | Level [2,3,4] |
| Brendel et al. [82] | ETHZ Central, TUD Crossing, i-Lids AB, UBC Hockey and ETHZ Soccer datasets | Maximum-weight independent set (MWIS) based tracker | Multi-Target Tracking (MTT) | Long-term occlusion handling | Level [2,3] |
| He et al. [83] | PETS09, TUD Statmitte, TUD Crossing and ETHMS datasets | Connected component model with MWIS | Multi-Target Tracking (MTT) | | Level [2,3] |
| Gaidon et al. [84] | KITTI Benchmark and PASCAL-to-KITTI dataset | Online Domain Adaptation for Multi-Object Tracking | Multi-Target Tracking (MTT) | Generic detector and video adaptation fro tracking | Level [2,3] |
| Ondruska et al. [85] | Simulated data | End-to-end recurrent neural network (RNN) tracker | Multi-Target Tracking (MTT) | No data association required | Level [2,3] |
| Dequaire et al. [86] | Real-world environment | End-to-end recurrent neural network (RNN) tracker | Multi-Target Tracking (MTT) | | Level [2,3] |
| Milan et al. [87] | MOTChallenge 2015 benchmark | Online recurrent neural network (RNN) tracker | Multi-Target Tracking (MTT) | | Level [2,3] |
| Ristani et al. [88] | Multi-cameras system data | Neural networks | Multi-Target Tracking (MTT) | Features learnt multi-cameras and Re-identification | Level [3, 4] |
| Liu et al. [89] | Multi-camera systems data | Generalized Maximum Multi-Clique optimization | Multi-Target Tracking (MTT) | | Level [2,3] |
| Park et al. [90] | Monocular camera data | 3D object tracking | Multi-Target Tracking (MTT) | 3D object Tracking for augmented reality applications | Level [2,3] |
| Scheidegger et al. [91] | Single camera data | Multi-Bernoulli mixture tracking filter | Multi-Target Tracking (MTT) | | Level [2,3,4] |
| Yan et al. [92] | Lidar data | 3D LIDAR based tracking with Support Vector Machine (SVM) classifier | Multi-Target Tracking (MTT) | Online classification of humans | Level [2,3] |
| Leal-taixe et al. [93] | Camera data | Interaction feature strings with Random Forest method | Multi-Target Tracking (MTT) | Scene understanding | Level [2,3] |
| Choi et al. [94] | Video data | Discriminative model | Multi-Target Tracking (MTT) | Group activity recognition | Level [2,3] |